# Automatic Colon Polyp Detection using Region based Deep CNN and Post Learning Approaches

Younghak Shin[1,2], Member, IEEE, Hemin Ali Qadir[2,3], Lars Aabakken[2,4], Jacob Bergsland[2] and Ilangko Balasingham[1,2], Senior Member, IEEE

[1]Department of Electronic Systems, Norwegian University of Science and Technology (NTNU), Trondheim, Norway
[2]Intervention Centre, Oslo University Hospital, Oslo Norway
[3]Department of Informatics at the University of Oslo (UiO), Oslo, Norway
[4]Department of Transplantation, Faculty of Medicine, University of Oslo, Norway

Corresponding author: Younghak Shin (e-mail: shinyh0919@gmail.com).

This work was supported by the European Research Consortium for Informatics and Mathematics (ERCIM) 'Alain Bensoussan' Fellowship Programme, Research Council of Norway through the MELODY project under the contract number 225885/O70 and Research Council of Norway through the industrial Ph.D. project under the contract number 271542/O30.

**ABSTRACT** Automatic detection of colonic polyps is still an unsolved problem due to the large variation of polyps in terms of shape, texture, size and color, and the existence of various polyp-like mimics during colonoscopy. In this study, we apply a recent region based convolutional neural network (CNN) approach for the automatic detection of polyps in images and videos obtained from colonoscopy examinations. We use a deep-CNN model (*Inception Resnet*) as a transfer learning scheme in the detection system. To overcome the polyp detection obstacles and the small number of polyp images, we examine image augmentation strategies for training deep networks. We further propose two efficient post-learning methods such as, automatic false positive learning and off-line learning, both of which can be incorporated with the region based detection system for reliable polyp detection. Using the large size of colonoscopy databases, experimental results demonstrate that the suggested detection systems show better performance compared to other systems in the literature. Furthermore, we show improved detection performance using the proposed post-learning schemes for colonoscopy videos.

## I. INTRODUCTION

Colorectal cancer (CRC) is the second most lethal cancer in the USA for both genders, causing 50,260 deaths in 2017 alone, a 2.18% increase from the previous year [1]. Most instances of CRC arise from growths of glandular tissue in the colonic mucosa known as adenomatous polyps. Mostly initially benign, some of these polyps become malignant over time, eventually leading to death, unless detected and treated appropriately. Therefore, the detection and removal of polyps in the early stage is an essential clinical procedure to prevent CRC [2].

Currently, colonoscopy represents the gold standard tool for colon screening. During a colonoscopy, clinicians inspect the intestinal wall in order to detect polyps. However, colonoscopy is an operator dependent procedure where the polyp miss-detection rate is about 25% [3]. The missed polyps can lead to a late diagnosis of CRC, in the worst case reducing survival rate to 10% [4]. Therefore, studies to develop computer-aided polyp detection are highly desirable.

Over the last two decades, various computer-aided detection (CAD) systems have been proposed to increase polyp detection rates [5]-[16]. In earlier studies, color, texture and shape based features such as color wavelet, local binary pattern (LBP) and edge detection were used to distinguish polyps from the normal mucosa [5][6][7]. However, these feature patterns are frequently similar between polyp and polyp-like normal structures, resulting in decreased performance. For more sophisticated detection, a valley information based Polyp appearance model has been suggested for polyp localization [8] and further improved versions with preprocessing methods for removing false positive regions have been proposed [9][10]. In [11][12], edge shape and context information were used to improve discriminative power between polyps and other polyp-like structures. To address balanced training between polyp and non-polyp images, an imbalanced learning scheme with a discriminative feature learning was proposed [17].

Recently, the region-based CNN approaches, *R-CNN* [18], *Fast R-CNN* [19] and *Faster R-CNN* [20] have shown considerable progress in object detection fields using natural image datasets. Unlike conventional hand-crafted feature based object detection approaches; *e.g.*, color wavelet, local binary pattern (LBP) and edge detection, the region-based CNN methods adopt the deep learning approach to learn rich feature representations automatically using deep-CNN architectures.

In the initial R-CNN study [18], external region proposal methods were adopted, such as Selective Search [21] and Edge Boxes [22], to train a CNN model (*e.g.*, *AlexNet*). However, each proposed region is needed to pass to the independent deep-CNN, resulting in a slow detection speed. To mitigate this problem, in the Fast R-CNN work [19], a single-stage CNN training was proposed by using a RoI (region of interest) pooling technique which substantially improved the detection speed. Finally, in the Faster R-CNN method, the authors proposed a region proposal network (RPN) to avoid the use of external time-consuming region proposal methods [20]. The RPN works within the deep CNN, sharing CNN features with the Fast R-CNN detector by the alternating training scheme. This method shows improved detection performance both in accuracy and time. Most recently, so-called Mask R-CNN method was proposed by the same group [23]. They extend the Faster R-CNN method for more challenging object segmentation task by adding a branch for predicting an object mask.

Due to the large variation of polyps in terms of shape, texture, size, and color, automatic polyp detection is still a challenging problem. In this study, we focus on the polyp detection task using the recent deep learning approach. The Faster R-CNN method shows excellent performance in large-scale general image datasets [20] and was successfully applied to other applications such as pedestrian detection [24][25] and face detection [26]. Despite this success, there have been no studies applying the region-based CNN approach to polyp detection. The main obstacle may be the paucity of available labeled colonoscopy datasets compared to natural image datasets. Motivated by this, we apply the Faster R-CNN based deep learning framework to the automatic polyp detection. To overcome limited training samples, we adopt a transfer learning scheme using a deep CNN model and examine proper image augmentation strategies. Furthermore, two post-learning schemes are suggested to improve polyp detection performance in colonoscopy videos.

### A. RELATED WORK
Recently, utilizing the success of deep learning in many image processing applications, a CNN based approach has been proposed for polyp detection [13][14]. In addition, in the recent polyp detection challenge, i.e., 2015 MICCAI challenge [27], several teams used CNN based end-to-end learning approaches. Above mentioned works focused on the conventional CNN based feature extraction and classification for the task of polyp detection. In [16], the authors proposed a 3D fully convolutional network approach to use time information with CNN features from the consecutive colonoscopy recording.

The concept of transfer learning schemes as a means of overcoming insufficient training samples, i.e., the use of pre-trained CNN by large-scale natural images, was successfully applied in different medical applications such as standard plane localization in ultrasound imaging [28], automatic interleaving between radiology reports and diagnostic CT and MRI images [29]. In [30], the performance of transfer learning on different CNN architectures (AlexNet and GoogLeNet) is evaluated in thoracic-abnormal lymph node detection and interstitial lung disease classification. N. Tajbakhsh *et al* [15], demonstrated that pre-trained CNN (AlexNet), with a proper fine-tuning approach, outperforms training from scratch in some medical applications including polyp detection.

It is generally known that the image augmentation is an efficient tool to increase the number of training samples. In the recent CNN based polyp detection tasks [15][16], simple augmentations were applied to increase the number of training samples. Authors of [15] used the upscaling, translating and flipping to the polyp patch images while in [16], rotating and translating were similarly adopted.

Some studies have applied post learning schemes for polyp detection. In [16], a time information based video specific online learning method was proposed and integrated with trained CNN. However, to train network online, additional learning time is needed (1.23 sec processing time per frame). In [31], AdaBoost learning strategy was suggested to train an initial classifier with new selected negative examples (FPs). This is a similar concept to our false positive (FP) learning scheme. The authors used the conventional image patch based hand-craft features such as LBP and Haar instead of CNN features. In this study, we provide the performance comparison between our method and [31] on the same 18 colonoscopy video dataset.

### B. CONTRIBUTIONS
Our main contribution is four-fold:

First, to the best of our knowledge, this work is the first study applying the region-based object detection scheme for the polyp detection application. Compared to previous transfer learning schemes in medical applications [15][30], we adopt the recent very deep CNN network, i.e., Inception Resnet, which shows the state of the art performance in the natural image domain and we evaluate the effect of this network as a transfer learning for a polyp detection task.

Second, we evaluate proper augmentation strategies for polyp detection by applying various types of augmentation such as rotating, scaling, shearing, blurring and brightening.

Third, we propose two post learning schemes: false positive (FP) learning and off-line learning. In the FP learning scheme, we suggest post training our detector system with automatically selected negative detection outputs (FPs) which are detected from normal colonoscopy videos. This scheme is effective to decrease many of the polyp-like false positives and therefore can be useful clinically. In the off-line learning scheme, we further improve the detection performance by using the video specific reliable polyp detection and post-training procedure.

Finally, from the large amount of experiments using public polyp image and video databases (total 28 videos), we demonstrate that our detection model shows improved detection performance compared to other recent CNN based studies in colonoscopy image dataset. In addition, the two proposed post-learning methods successfully work for polyp detection in the colonoscopy video databases.

The remainder of this paper is organized as follows. In Section II, the proposed detection systems and methodological steps are introduced. In Section III, experimental datasets used in this study are described. In Section IV, evaluation metrics, experimental results and discussions are presented. Finally, we conclude this study in Section V.

## II. METHODS

In this section, we aim to introduce our proposed polyp detection system. Fig. 1 shows the entire polyp detection procedure. The first step for training the detector system is to perform an augmentation on the images in order to increase the number of useful polyp training samples. Next, region proposal network (RPN) proposes rectangular shaped regions that may include a polyp. In the Detector part, using the proposed regions in RPN, polyp classification and region regression are performed to predict final polyp region.

Finally, we propose further post-learning schemes, i.e., false positive (FP) and off-line learning, to improve polyp detection performance. We explain the details of each step in the following subsections.

### A. IMAGE AUGMENTATION

For a stable training of deep-CNN models, normally a large amount of training dataset is needed, *e.g.*, AlexNet is trained on 1.2 million of ImageNet dataset [32]. However, obtaining a large number of polyp images with the corresponding ground truth of polyp masks is generally quite difficult. To overcome this lack of images, image augmentation, such as rotating and flipping of the originals, increases the number of training samples. However, this augmentation strategy needs to be carefully applied based on an adequate understanding of the application domain. In other words, the augmentation should be generated by considering real colonoscopy images and have enough variations to avoid overfitting. In this study, we aim to evaluate different augmentation strategies for the deep-CNN based polyp detection system.

In colonoscopy recordings, polyps show large variation in scale, location and color. In addition, changing camera viewpoints and lighting conditions lead to varying image definition and brightness. Therefore, we consider not only simple rotating and flipping but also zooming, shearing, blurring and altering brightness as polyp image augmentation strategies.

Fig. 2 shows an example of 9 different image augmentations performed on one polyp image for use in the training of our detection system. We rotate the image clockwise 90, 180, and 270 degrees. We also use horizontal and vertical flipping. To create different scales of polyp images, *e.g.,* Fig. 2-(d) and (e), we perform zoom-in and out with specific zooming parameters; *i.e.,* 10% and 30% zoom-in.

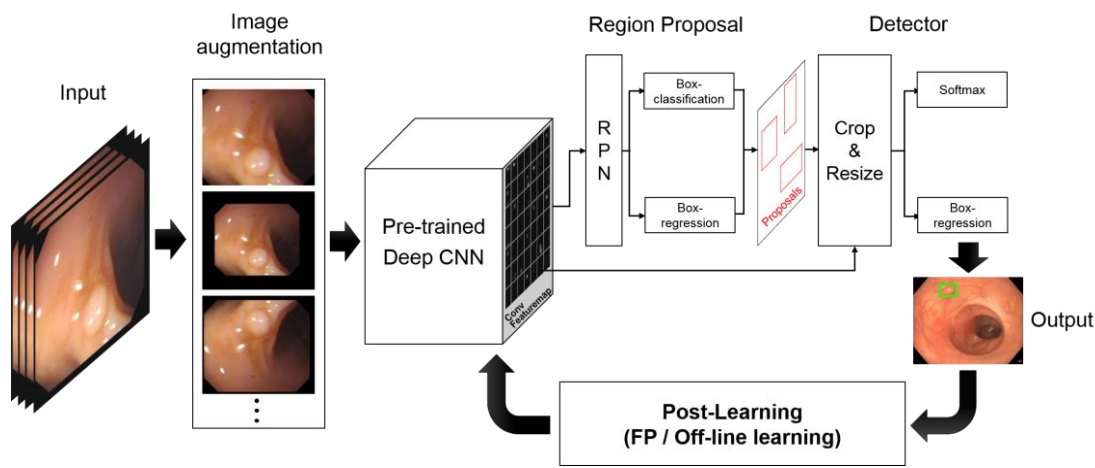

**FIGURE 1.** Proposed polyp detection system. The detector system consists of three main part, region proposal network, detector and post-learning. For training the detector system, domain specific image augmentation and transfer learning using pre-trained deep CNN are adopted.

We perform four different shearing operations: two along the x-axis to shear the images from left to right and two along the y-axis to shear them from top to bottom. For blurring the image in Fig. 2-(b), we apply Gaussian filtering with specific standard deviation parameters. Finally, brightness control, *e.g.*, Fig. 2-(f) and (g), is performed by adjusting the image intensity using the specific contrast limit for generating bright and dark images.

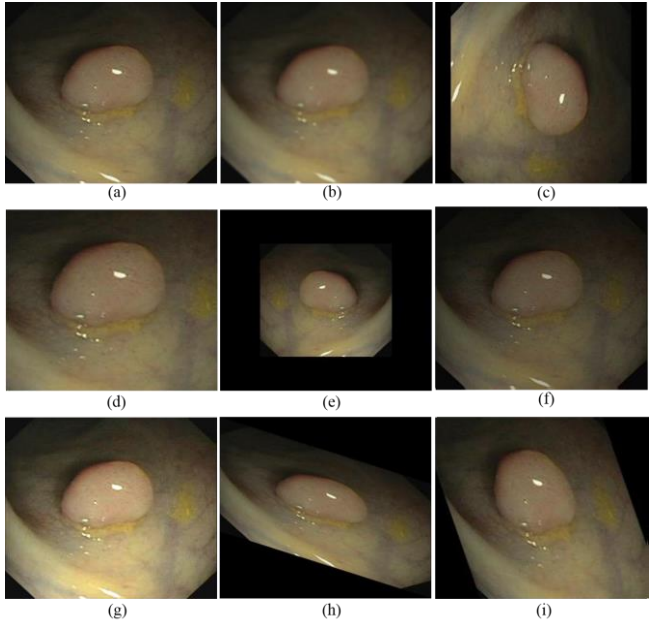

FIGURE 2. Example of polyp image augmentation. (a) original polyp image frame, (b) blurred image with 1.0 of standard deviation, (c) 90 degree rotated image, (d) 10% zoom-in image, (e) 30% zoom-out image, (f) dark image, (g) bright image, (h) sheared image by y-axis, (i) sheared image by x-axis.

Using the above mentioned augmentations, we design four different augmentation strategies to compare the augmentation effect of the deep-CNN based polyp detection in polyp image and video databases. First, for training the detector system, we use only original images without any augmentation (w/o augmentation). Second, we apply three rotations of 90, 180, 270 degrees and horizontal/vertical flips to the original images (Rot-augmentation).

Third, for Augmentation-I, we aim to consider more different shapes of polyps. Therefore, we apply four different types of shearing to each original image. Furthermore, three zoom-out (10, 30 and 50%) and one zoom-in (10%) augmentations are applied to the original images and the three rotated and flipped images. Because detection of small size polyps within image frames is much more difficult than that of large size polyps, we apply imbalanced zooming, i.e., three zoom-out and one zoom-in. For those polyps located near the four corners of the image frames, much of the polyp can disappear after the zoom-in process, as a result, these augmented images are excluded for training our model. The total number of training images after applying the Augmentation-I (Aug-I) is 18594.

Lastly, for Augmentation-II (Aug-II), we further consider different resolution and brightness of the colonoscopy image frame. This might be helpful for polyp video detection with different variations of frames. We adopt all augmented images used in Aug-I, and add one final augmentation consisting of blurring, brightening and darkening the original, three rotated and two flipped images. In this way, we generate 28600 images, producing the largest augmented training dataset.

Note that for the parameters of zooming, blurring and brightness augmentations could be changed a bit depending on the resolution and brightness of the original image and minimum and maximum polyp size of each polyp image frame.

### B. REGION PROPOSAL METHOD
In this study, we adopt the region proposal network (RPN) which was introduced in the Faster R-CNN method [19] to obtain polyp candidate regions in polyp frames.

Here, we briefly introduce how the RPN method works. The RPN takes any size of input images and outputs a number of rectangular shaped region proposals, each with an objectness score. Each region is expressed by (x, y, w, h), where x, y is the object position of the top-left corner and w, h represents the width and height of the object. The input training image is passed by the pre-trained deep-CNN as shown in Fig. 1. This network can be trained from scratch or pre-trained by a large-scale dataset. Usually, the feature map of the last convolutional layer in the whole network (e.g., conv5 layer on VGG network in [20]) is used for the RPN.

The RPN slides a $3 \times 3$ window on the feature map. Then, each sliding window is mapped into a fixed size feature vector followed by two sibling $1 \times 1$ fully connected layers; i.e., a box-regression layer to predict location (x, y, w, h) of proposals and a box-classification layer to predict object (polyp and background) scores (please see Figure 3 of [20] for details). At the center of each sliding window, k reference boxes (anchor boxes) are generated to make the system less sensitive to changes in the shape of objects. The fixed k=9 anchor boxes with three different scales and aspect ratios are used in the original paper [20]. However, in this study, we use k=12 with four scales [0.25, 0.5, 1.0, 2.0] and three aspect ratios [0.5, 1.0, 2.0] to consider larger variations of polyps. For each k proposal, RPN predicts the locations and class scores.

### C. FAST R-CNN DETECTOR
The second module of the Faster R-CNN is object detector which was introduced in the Fast R-CNN work [19]. As shown in Fig. 1, the inputs of the detector are image frame and corresponding region proposals obtained from the previous RPN step. The input image frame is passed by several convolutional and pooling layers of the deep-CNN to produce feature map of the last convolution layer. Then, each region proposal which is also called the region of interest (RoI) is sent to a RoI pooling layer to generate a fixed-size feature vector from the feature map.

Note that for different sized RoIs, the same fixed-size feature vector is needed because the following fully connected layer, adopted from a pre-trained network expects the same size input [19]; and, in the RoI pooling layer, each rectangular region expressed by height ($h$) and width ($w$) is projected onto the feature map. Then, simply max-pooling is executed to generate a fixed size feature vector; *i.e.*, $h \times w$ region proposal is max pooled using a sub-window of size $h/H \times w/W$, where, $H$ and $W$ are network model dependent fixed parameters; *i.e.*, it should be compatible with the first fully connected layer of the model. In this study, we use the Tensorflow framework to implement a Faster R-CNN, where instead of using the RoI pooling layer, 'crop and resize' operation which was recently adopted in [33][34]. This operation utilizes the bi-linear interpolation to make the same purpose fixed size feature vector. And then, each vector is fed into two sibling layers, a softmax layer and a box regression layer, the former to estimate class score and the latter to refine the proposal coordinates.

### D. IMAGE TRANSFER LEARNING WITH PRE-TRAINED DEEP CNN

Transfer learning is an efficient technique for applying a deep learning approach to many applications [15][18][30]. It is especially advantageous for training when there is a paucity of available labeled training data. To apply the transfer learning scheme, we utilize a pre-trained network trained by large-scale natural images. Then, we aim to fine-tune our detection system with the available polyp training dataset.

For a CNN network, we consider a recent deep-CNN model, *i.e.*, '*Inception Resnet*' [35]. The Inception Resnet shows the state-of-the-art classification performance in many different challenging datasets [35] and also in object detection tasks [33]. This network combines the advantages of both recent *Resnet,* [36], i.e., residual learning: adding residual connections between stacked layers to obtain optimization benefit, and *Inception* [37][38] networks, i.e., inception module: design parallel paths of convolution with different receptive field sizes to capture various types of features. In the Inception Resnet the combined Inception-Resnet modules (Inception-Resnet-A, B and C in [35]) were used for the efficient training of a deep network. Each Inception-Resnet module is repeated several times, with the total depth of the network being over 100 layers. Two versions of Inception Resnet have been introduced in [35] and we use a deeper version called Inception Resnet-v2. More detailed information about the network architecture and implementation is available in [35][39].

The deep-CNN network that we use for initializing our detector network was pre-trained on Microsoft's (MS) COCO (Common Objects in Context) dataset [40]. This dataset is well known for having a large number of object instances per image as compared to other large-scale datasets such as ImageNet and PASCAL [20][40]. For training of the deep-CNN, 112K images (*i.e.*, 80K of '2014 train' and 32K of '2014 val' images [33]) were used. This training dataset contains 90 different common object categories such as a people, bicycles, dogs, cars *etc*.

### E. TRAINING DETECTOR

In the initial Faster R-CNN work [20], the RPN and the Fast R-CNN detector were trained by sharing CNN features via a 4-step alternating training scheme. Later, more efficient end-to-end joint training was suggested by the same authors, and used in Tensorflow implementation for a Faster R-CNN [33]. For the fine-tuning of the detector systems, trained weights of the pre-trained model are used for initial weights and all weights of new layers for the RPN and the Fast R-CNN detector are randomly initialized.

For training of RPN, the positive and negative training samples should be selected from the anchor boxes by computing IoU (Intersection-over-Union) with the ground truth of the object location. In the Faster R-CNN work [20], 0.3 and 0.7 IoU values were adopted. Specifically, when the anchor has an IoU overlap higher than 0.7 with the ground truth location, a positive label is assigned. A negative label is assigned when the IoU overlap is lower than 0.3. However, this value may not be optimal for polyp detection tasks. In this study, we compare the detection performance of different IoU values and we choose 0.3 and 0.6 for selection of negative and positive training samples. We include this comparison results in Table II. As used in [20][33], to avoid the high overlap of proposals and detection output, non-maximum suppression (NMS) is adopted with 0.7 of IoU for training and 0.6 of IoU for testing. For each image frame, the maximum number of proposals is set to 300.

We use the stochastic gradient descent (SGD) method with a momentum of 0.9 [32], as used in the Faster R-CNN work [20]. In each iteration of the RPN training, 256 training samples are randomly selected from each training image where the ratio between positive ('polyp') and negative ('background') samples is 1:1. We set the maximum number of epochs to 30 with the learning rate equal to 1e-3.

### F. FALSE POSITIVE LEARNING

For reliable polyp detection supporting tools, the small FP, i.e., false alarms, is desirable from clinical point of view. However, in polyp detection task, existence of the polyp-like false positives (FPs) is a major difficulty. More specifically, in a colonoscopy video recording, some parts closely resemble polyp characteristics such as, circle shaped light reflections, and overexposed regions, intestinal contents and black hole parts from luminal regions [27] and these would be incorrectly detected as polyps. These FPs result in performance degradation (especially in precision) in colonoscopy video detection.

In this study, we use the publicly available CVC-CLINIC dataset to train the detector system. In this dataset, only 612 image frames with polyps and corresponding polyp positions are provided. As we mentioned in Section II-E, the detector system is trained with the polyp objects (*i.e.,* positive samples) based on the annotated ground truth of polyp masks and specified IoU values. The negative samples for training (*i.e.,* normal background regions) are randomly selected within the polyp image frames. It is difficult to have exact bounding boxes around the polyp-like mimics for the randomly selected negative samples. Therefore, the detector system, which is only trained with the polyp images, tends to have many polyp-like FPs when testing the colonoscopy videos.

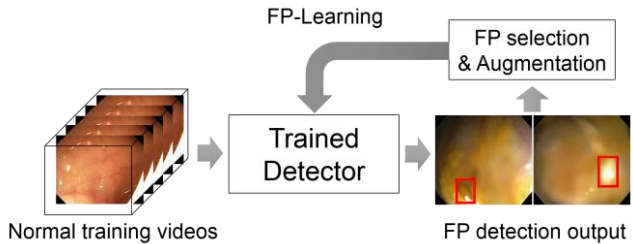

**FIGURE 3.** Procedure of proposed false positive learning scheme

To overcome this problem, we propose an automatic FP learning scheme in order to make a more robust detection system. Fig. 3 illustrates the proposed procedure for automatic FP learning. We use the 5 normal videos from 10 ASU-Mayo normal video dataset (see Section III) to collect detected polyp-like FPs. Note that any annotated training dataset (e.g., polyp images frames, polyp videos and normal videos) can be used for collecting polyp-like FPs. Using the initial detector system trained by the 612 polyp images, we first test these 5 videos in order to collect polyp-like FPs with the corresponding bounding box locations (x, y, w, h). Among the collected FPs, we only select strong FPs which have high polyp-scores, i.e., we use class-score information from the detector system. Then, the initial detector system is re-trained with the selected polyp-like FPs and corresponding bounding boxes.

In this study, we set the 99% of score threshold to select FP detections commonly considered as a polyp from the different normal colonoscopy videos. If we set a smaller score threshold, then there will be a large variation in FP detections and it would make it difficult to train the detector system. After collecting the FPs, we apply the image augmentation to increase the number of training samples.

For image augmentation of selected FPs, 5 rotations of the original images are applied. This is because the polyp-like FPs and corresponding bounding boxes are automatically detected by the previous detector system and they have high polyp-scores. We expect that the re-trained system will be robust in reducing the number of FPs after this FP learning process, which efficiently increasing the detection precision.

Fig. 4 shows several examples of the selected FP images from the 5 normal training videos. The FPs have features similar to real polyps, with over 99% on the polyp-score. 654 FP images and bounding boxes are automatically collected, and after augmentation 3922 images and bounding boxes are used for FP learning.

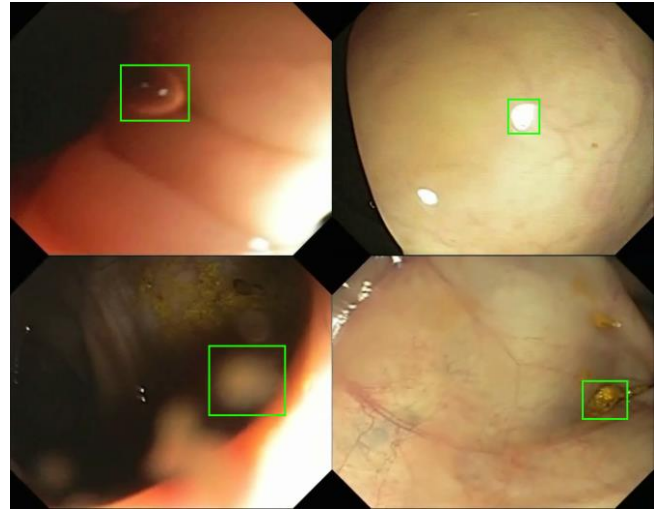

**FIGURE 4.** Example of automatically selected FP regions (represented by green box). Upper left: circle shaped water bubble, Upper right: circle shaped light reflection, Bottom left: circle shaped reflection from camera, Bottom right: intestinal content.

### G. OFF-LINE LEARNING FOR VIDEO DETECTION

Even though transfer learning and image augmentation techniques are applied to the detection systems, it is still challenging to obtain high detection performance in some colonoscopy videos due to: large variation of polyps with respect to scale and location; variable camera viewpoints and lighting conditions. In addition, each colonoscopy video has different types of FPs. Therefore, it is quite difficult to improve performance given the limited training dataset.

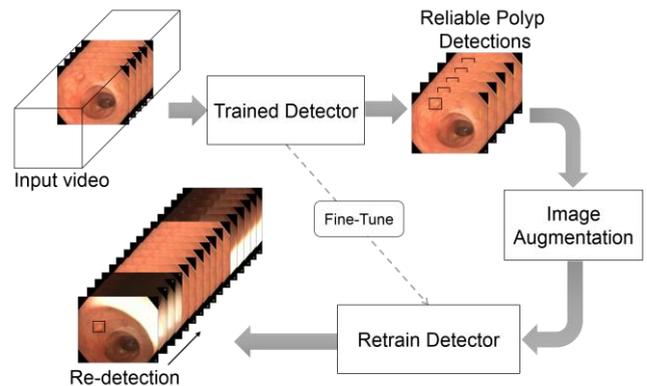

**FIGURE 5.** Procedure of the post off-line learning scheme

In this section, we propose a simple video-specific post learning process for the purpose of off-line analysis of each colonoscopy video. Fig. 5 illustrates the proposed off-line learning procedure. We use our detector system trained by the initial training dataset (Aug-I) for the reliable detection of new polyp regions in each test video. On each video, we first run the Aug-I model to collect reliable polyp regions and automatically generate polyp masks (ground-truth) for the detected regions. Secondly, we apply augmentations to the collected polyp regions and the corresponding polyp masks. We retrain the detector system using those collected polyps from the video being tested. Finally, we test the video again using the new trained detector system. We define this framework as offline learning process because the model is retrained after the entire video is tested not while it is being tested (online-learning). We expect that after this video-specific off-line learning process it will be possible to detect larger variations of polyps in each video. At the same time, the detector can learn video specific FPs.

## III. EXPERIMENTAL DATASETS

In this study, we use publicly available polyp-frame datasets, CVC-CLINIC [10] and ETIS-LARIB [41], and two colonoscopy video databases, ASU-Mayo Clinic Colonoscopy Video dataset [12] and CVC-ClinicVideoDB dataset [31]. These datasets were used in the recent challenge 'Endoscopic Vision Challenge' in MICCAI (Medical Image Computing and Computer Assisted Intervention) 2015 conference [27].

The CVC-CLINIC dataset contains 612 polyp image frames with a pixel resolution of $388 \times 284$ pixels in SD (standard definition). All images were extracted from 31 different colonoscopy videos which contain 31 unique polyps. The ETIS-LARIB dataset comprises 196 polyp images which are generated from 34 colonoscopy videos. Each image has an HD (high definition) resolution of $1225 \times 966$ pixels. This dataset contains 44 different polyps with various sizes and appearances. At least one polyp existed in all 196 images, with the total number of polyps being 208. All ground truths of polyp regions for both datasets were annotated (*e.g.*, see Fig. 6) by skilled video endoscopists from the corresponding associated clinical institutions. Both CVC-CLINIC and ETIS-LARIB polyp-frame datasets were used for the polyp localization challenge [27]. In this study, for a fair comparison of detection performance with the challenge results, we follow the same evaluation strategy used in the challenge, *i.e.*, 612 images from the CVC-CLINIC dataset were used for the training of detection systems and 196 images from the ETIS-LARIB dataset were used for evaluation.

For the evaluation of polyp detection in colonoscopy videos, we use two different video databases. The ASU-Mayo Clinic Colonoscopy Video dataset contains 20 training and 18 testing videos. Due to license problems the ground truth of the test set is not available. Therefore, in this study, we use only the 20 training videos for the evaluation of proposed detection schemes. These 20 videos consist of 10 positive and 10 negative videos; *i.e.*, positive videos include some polyp image frames and negative videos are normal frames with no polyps. In 10 positive videos, there are a total of 5402 frames with a total of 3856 polyp frames. In 10 negative videos, there are 13500 frames without polyps. Each frame of the video database comes with a binary ground truth in which each polyp is annotated by clinical experts. Each positive video includes a unique polyp. Within each video, there is a large degree of variation with respect to scale, location and brightness. In addition, some polyp frames include artifacts such as tools for water insertion and polyp removal.

The recent CVC-ClinicVideoDB video dataset comprises 18 different SD videos of different polyps. In this dataset, 9221 frames out of 10924 frames contain a polyp, and the size of the frames is $768 \times 576$. Each frame of the video databases comes with a binary ground truth, in which each polyp is annotated by clinical experts. Each positive video includes a unique polyp. Within each video, there is a large degree of variation with respect to scale, location and brightness. In addition, some polyp frames include artifacts such as tools for water insertion and polyp removal.

We use both the ASU-Mayo Clinic and the ClinicVideoDB Colonoscopy Video databases to examine the overall polyp detection performance of the model that was trained by the 612 images of CVC-CLINIC dataset. In case of ASU-Mayo Clinic dataset, we use the 10 positive and 5 negative videos for testing the detection systems. For evaluation of the proposed FP learning scheme, which is explained in Section II-F, we use the remaining 5 negative videos to collect some normal images and then retrain the trained model with the collected normal parts.

## IV. RESULTS AND DISCUSSION

### A. EVALUATION METRICS

In the context of this study, we use the term "polyp detection" as the ability of the model to provide the location of the polyp within a given image. We use the same evaluation metrics presented in the MICCAI 2015 challenge [27] to perform fair evaluation of our polyp detector performance and benchmark our results with the results from the challenge. Since the output of our model is the four rectangular shaped coordinates (x, y, w, h) of the detected bounding box, we define the following parameters as follows:

**True Positive (TP):** correct detection output if the detected centroid falls within the polyp ground truth.
**False Positive (FP):** any detection output in which the detected centroid falls outside the polyp ground truth.
**False Negative (FN):** polyp is missed in a frame containing a polyp.
**True Negative (TN):** no detection output at all for negative (without polyp) images.

Note that if there is more than one detection output, only one TP is counted per polyp. Based on the above parameters, the three usual performance metrics, i.e., precision (pre), recall (rec) and specificity (spe) can be defined:

$$Pre = \frac{TP}{TP+FP}, Rec = \frac{TP}{TP+FN}, Spe = \frac{TN}{FP+TN} \quad (1)$$

Furthermore, to consider balance between precision and recall we also use *F1* and *F2* scores which are:

$$F1 = \frac{2 \times Pre \times Rec}{Pre + Rec}, F2 = \frac{5 \times Pre \times Rec}{4 \times Pre + Rec} \quad (2)$$

We further include following metrics to evaluate performance of polyp detection performance in colonoscopy videos [31]:

**Polyp Detection Rate (PDR):** measure to know if a method can find the polyp at least once (100%) or not (0%) in a sequence of polyp video frames.

**Mean Processing Time per Frame (MPT):** It is the actual detection processing time taken by a method to process a frame and display the detection result.

**Reaction Time (RT):** Defines how fast a method reacts when a polyp appears in a sequence of video frames. It can be compute in two ways as follows:

**in frames:** It calculates the delay in frame between first TP detection and first appearance of the polyp in a sequence.

**in seconds:** Considering 25fps, it calculates the delay in seconds between first TP detection and first appearance of the polyp in a sequence.

### B. EVALUATION OF POLYP FRAMES

In this section, we report the performance of our polyp detection system, trained with 612 CVC-CLINIC dataset on still frame images using the 196 ETIS-LARIB dataset. Table I shows the evaluation results for the four different image augmentation strategies utilized.

TABLE I
COMPARISON OF POLYP FRAME DETECTION RESULTS USING FOUR DIFFERENT AUGMENTATION STRATEGIES

| Training dataset | TP | FP | FN | Pre (%) | Rec (%) | F1 (%) | F2 (%) |
|---|---|---|---|---|---|---|---|
| w/o augmentation | 82 | 89 | 126 | 48 | 39.4 | 43.3 | 40.9 |
| Rot-augmentation | 147 | 99 | 61 | 59.8 | 70.7 | 64.8 | 68.2 |
| Augmentation-I | 167 | 26 | 41 | 86.5 | **80.3** | **83.3** | **81.5** |
| Augmentation-II | 148 | 14 | 60 | **91.4** | 71.2 | 80 | 74.5 |

The results presented in Table I show that when the detector model is trained with a large number of training images such as Aug-I and –II, it shows better detection performance than that trained with a small number of images. This means that having a large enough training sample with more variation leads to performance improvement. However, even though the detector model with Aug-II has a much larger number of training images (28600 images) than the Aug-I (18594 images), the Aug-I shows better detection performance in terms of recall, F1 and F2 scores.

In Fig. 6, we investigate some testing polyp frames from the ETIS-LARIB dataset which are not correctly localized by Aug-II but are correctly localized by Aug-I. The polyps in these three frames are very difficult to see via the naked eye. The second row shows that all but one polyp from the first column is successfully detected based on Aug-I, while the detector system based on Aug-II did not detect any polyps at all (see the third row).

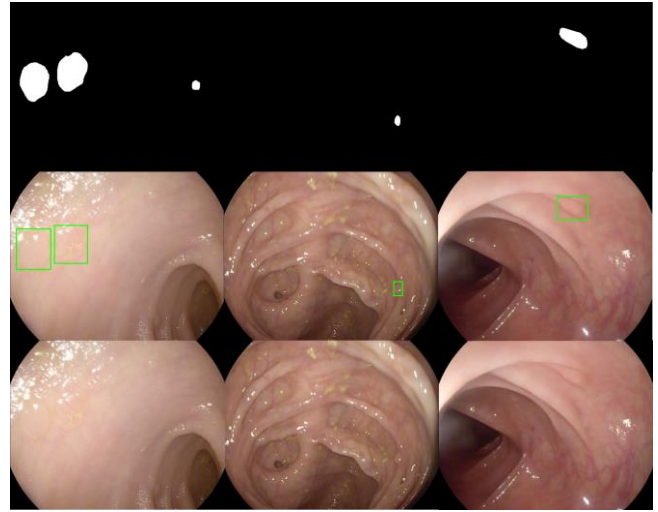

**FIGURE 6.** Detection examples of difficult polyps in ETIS-LARIB test images. The first row shows the ground truth images of the test images below. The second and third rows represent detection results from Augmentation-I and Augmentation-II respectively.

We surmise that the reason is that we apply image augmentation consisting of additional blurring, brightening and darkening into the low definition training dataset during Aug-II to detect polyps in the high definition test dataset. Such augmentation methods can have a detrimental effect on image quality, making it more difficult to form clear polyp features during the training stage. This results in difficulty detecting unclear polyps as shown in Fig. 6, as well as resulting in much less TP (148) compared with the Aug-I (167) in Table I. Perhaps other augmentation strategies will improve detection performance. We note that it is important to fully consider domain-specific characteristics as well as the image quality of the training and test dataset when applying augmentation to increase the number of training samples.

In this study, we use a transfer learning scheme with a pre-trained deep-CNN model, *i.e.*, Inception Resnet trained by MS COCO dataset (Section III-D). For all results in Table I, the pre-trained model was applied and then we fine-tune the model with specific augmentation strategies. We evaluate our best detection model (Aug-I) using the concept of training from scratch [15]; *i.e.,* Inception Resnet is randomly initialized and trained with only Aug-I training images. In this case, we obtain very poor detection results, *i.e.,* 33.7% of recall and 27.1% of precision, compared with the transfer learning based model. The poor results are related to the number of original training images. We only have 612 images, which are not enough to extract rich features from such a deep-CNN model even after applying our augmentations.

As mentioned in Section II-E, in Table II, we compare detection performance of different IoU values for selection of positive and negative training samples. We use the Aug-I training images to train a detector system with different combinations of IoU values represented in Table II. The results show that there is no perfect winner in all performance metrics, and the performance difference is not large among different IoU selections. We use the 0.6 and 0.3 IoU values in this study since these values show the smallest number of FP.

TABLE II
COMPARISON OF POLYP FRAME DETECTION RESULTS USING FOUR DIFFERENT IoU COMBINATIONS

| IoU (Positive, Negative) | TP | FP | FN | Pre (%) | Rec (%) | F1 (%) | F2 (%) |
|---|---|---|---|---|---|---|---|
| 0.7, 0.3 | 157 | 34 | 51 | 82.2 | 75.5 | 78.7 | 76.7 |
| 0.6, 0.4 | 163 | 31 | 45 | 84.0 | 78.4 | 81.1 | 79.4 |
| 0.7, 0.4 | 171 | 37 | 37 | 82.2 | **82.2** | 82.2 | **82.2** |
| 0.6, 0.3 | 167 | 26 | 41 | **86.5** | 80.3 | **83.3** | 81.5 |

Fig. 7 illustrates channel activations of a specific CNN layer after training the Aug-I based polyp detection model. To fairly examine polyp activations, we choose a polyp image (*upper left* in Fig. 7) which is correctly detected by the Aug-I model with 100% class score. Then, we visualize activations on 192 convolutional channels (*upper right* in Fig. 7) at 1×1 convolutional layer of Inception-Resnet-B module in Inception Resnet [35]. The bright parts (white pixels) represent strong activations corresponding to the same position in the original test image [30]. As we can see in the *upper right* of the figure, many different channels have strong activations at the polyp position in the test image.

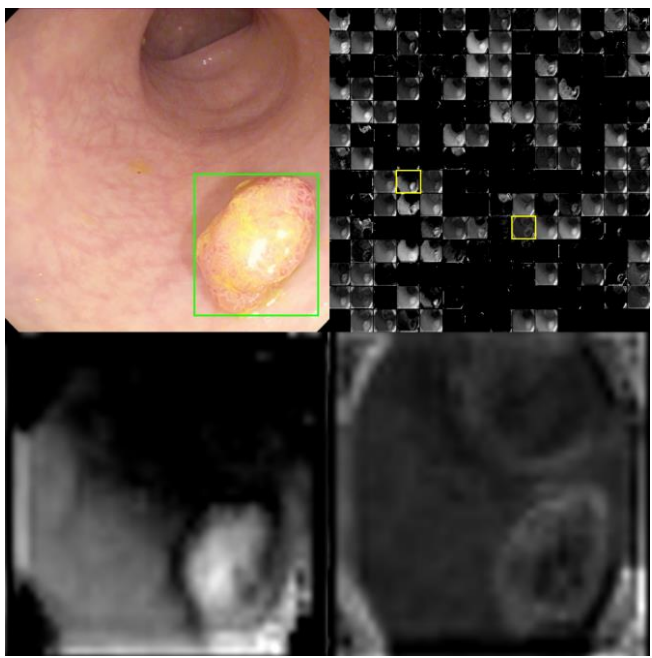

FIGURE 7. Visualization of CNN channel activations for a test image after training the Aug-I based detection model. Upper left: detection output for the test image, Upper right: Activations on all convolutional channels (192) at 1×1 convolutional layer in Inception-Resnet-B module, Bottom left and right: Activation map for the specific channel indicated by the left and right yellow box at the upper right figure.

More specifically, we emphasize two specific channels as shown in the *bottom left* and *right* of Fig. 7. These two activations correspond to the left and right yellow boxes at the *upper right* of the figure. We observe that the channel in the *bottom left* has strong activations inside the polyp part. On the other hand, the *bottom right* channel activates on edges of the polyp. This means that both channels extract polyp features efficiently and may contribute to polyp detection with high score.

### C. COMPARISON WITH OTHER METHODS

In Table III, we compare the detection performance of our model with the results of those of other teams in the 2015 MICCAI challenge [27] in which the exact same dataset was used. We include the top three results from each team: CUMED, OUS and UNS-UCLAN. All three teams used CNN based end-to-end learning for the polyp detection task. CUMED employed a CNN based segmentation strategy [41] where pixel-wise classification was performed with ground-truth polyp masks. The OUS team adopted the AlexNet CNN model [32] along with the traditional sliding window approach for patch-based classification [27]. The UNS-UCLAN team utilized three CNNs for feature extraction of different spatial scales and adopted one independent Multi-Layer Perceptron (MLP) network for classification [27]. We also include the combined detection performance from the top two teams (CUMED & OUS) which was presented in Table VI of [27].

TABLE III
COMPARISON OF POLYP FRAME DETECTION RESULTS WITH OTHER STUDIES

| Method | TP | FP | FN | Pre (%) | Rec (%) | F1 (%) | F2 (%) |
|---|---|---|---|---|---|---|---|
| CUMED | 144 | 55 | 64 | 72.3 | 69.2 | 70.7 | 69.8 |
| OUS | 131 | 57 | 77 | 69.7 | 63.0 | 66.1 | 64.2 |
| UNS-UCLAN | 110 | 226 | 98 | 32.7 | 52.8 | 40.4 | 47.1 |
| CUMED+OUS | 159 | 38 | 49 | 80.7 | 76.4 | 78.5 | 77.2 |
| Our model (Aug-I) | 167 | 26 | 41 | 86.5 | **80.3** | **83.3** | **81.5** |
| Our model (Aug-II) | 148 | 14 | 60 | **91.4** | 71.2 | 80 | 74.5 |

As can be seen in Table III, the results of our detection models based on Aug-I and -II are better than the results of each team in terms of all performance metrics: precision, recall, F1 and F2 scores. Specifically, our Aug-I model achieved a much larger TP, correctly detecting a total of 167 polyps out of a total 208 polyps in the ETIS-LARIB dataset, and with a smaller FP compared to all other teams.

Furthermore, our best model outperforms on all performance metrics the combined two best teams (CUMED & OUS). This means that the Faster R-CNN method, with the appropriate augmentation strategies, is very promising for polyp detection tasks compared to other CNN based methods.

Due to the use of different computer systems (mainly affected by GPU in deep learning), it is difficult to compare detection processing time directly. In this study, for testing of detection processing time, we use a standard PC with a NVIDIA GeForce GTX1080 GPU. We compute the detection processing time for each test image frame and average over all test images. The mean detection processing time (MPT) is about 0.39 sec per frame. Based on Table I of [27], The OUS and UNS-UCLAN have the same 5 sec processing time per frame, and the CUMED has 0.2 sec in NVIDIA GeForce GTX TITAN X GPU. Since we use a recent deep CNN model in our system detection times are not very fast. However, it is comparable with other CNN based polyp detection systems.

### D. EVALUATION OF COLONOSCOPY VIDEOS-I

In this section, we first evaluate the performance of the four different augmentation strategies on colonoscopy videos, in which polyp and normal mucosa (without polyp) frames are included. We evaluate 10 ASU-Mayo positive videos where one unique polyp is included with various changes in each video (see Section III). Table IV presents the results of the four augmentation strategies on the 10 positive videos. Again, Aug-I and –II show much better improvement in all performance metrics compared to the smaller number of training samples (i.e., w/o and Rot-augmentation). Consistently with Table I, Aug-I shows the larger number of TP (which results in a better recall) than Aug-II, while Aug-II has the smaller number of FP (which results in a better precision) than Aug-I.

TABLE IV
COMPARISON OF POLYP DETECTION RESULTS USING FOUR DIFFERENT AUGMENTATION STRATEGIES ON 10 AUS-MAYO POSITIVE VIDEOS

| Train Dataset | TP | FP | FN | TN | Pre (%) | Rec (%) | F1 (%) | F2 (%) |
|---|---|---|---|---|---|---|---|---|
| w/o aug | 1522 | 1246 | 2334 | 1105 | 55 | 39.5 | 46 | 41.8 |
| Rot-aug | 2343 | 1758 | 1513 | 824 | 57.1 | 60.8 | 58.9 | 60 |
| Aug-I | 3137 | 1145 | 719 | 769 | 73.3 | **81.4** | 77.1 | **79.6** |
| Aug-II | 2899 | 595 | 957 | 1131 | **83** | 75.2 | **78.9** | 76.6 |

Fig. 8 shows correct polyp detection from 10 different videos. The model used is based on Aug-I, the same as used in Table III. It successfully detects all 10 different types of polyps.

More specifically, in Table V, we compare the detection results of all the frames from all 10 videos using three different models: our best model based on Aug-I; and two proposed post learning methods, automatic FP learning and off-line learning, the latter both trained with the trained model based on Aug-I.

TABLE V
POLYP DETECTION RESULTS FOR 10 POSITIVE VIDEOS (EACH VIDEO HAS AT LEAST ONE POLYP FRAME)

| Method | TP | FP | FN | TN | Pre (%) | Rec (%) | F1 (%) | F2 (%) | PDR | RT (frames, sec) |
|---|---|---|---|---|---|---|---|---|---|---|
| Aug-I | 3137 | 1145 | 719 | 769 | 73.3 | 81.4 | 77.1 | 79.6 | 100% | 5.7, 0.22 |
| FP learning | 3008 | 412 | 848 | 1255 | **88** | 78 | 82.7 | 79.8 | 100% | 6, 0.24 |
| Off-line learning | 3245 | 677 | 611 | 1098 | 82.7 | **84.2** | **83.4** | **83.9** | 100% | 10.7, 0.428 |

In 10 positive videos, the total number of polyp and normal image frames is 3856 and 1546 respectively. The model based on Aug-I can correctly detect 3137 polyps out of 3856 with 1145 FPs, resulting in a recall of 81.4% but a precision of 73.3%. Compared to the still frame test results in Table III, the recall is similar (the difference is just 0.9%) but the precision is much degraded (13%) for the same model.

One reason for the high FP rate of this result is that some polyps did not clearly appear as shown in the first column of Fig. 9 and therefore were not annotated by experts. We noticed these missed annotated polyps based on the ground truth of constitutive frames in each video. However, our detection model Aug-I did in fact detect these missed polyps as shown in the second column of Fig. 9. Since they were not originally annotated, the detections of Aug-I model are considered as FPs for these frames in Table IV and V.

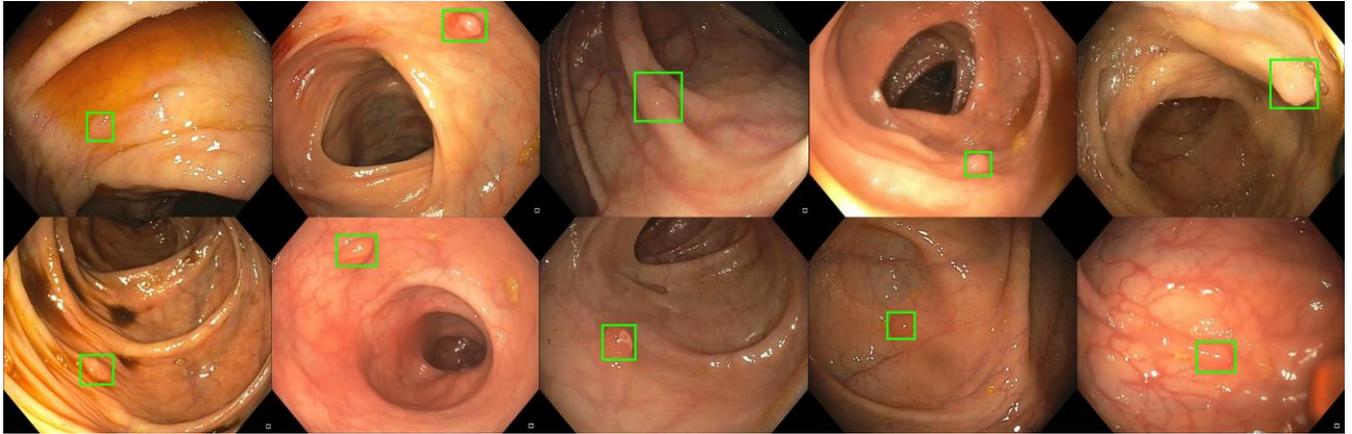

**FIGURE 8.** Example of correct polyp detections in 10 positive videos using the Augmentation-I trained detector model

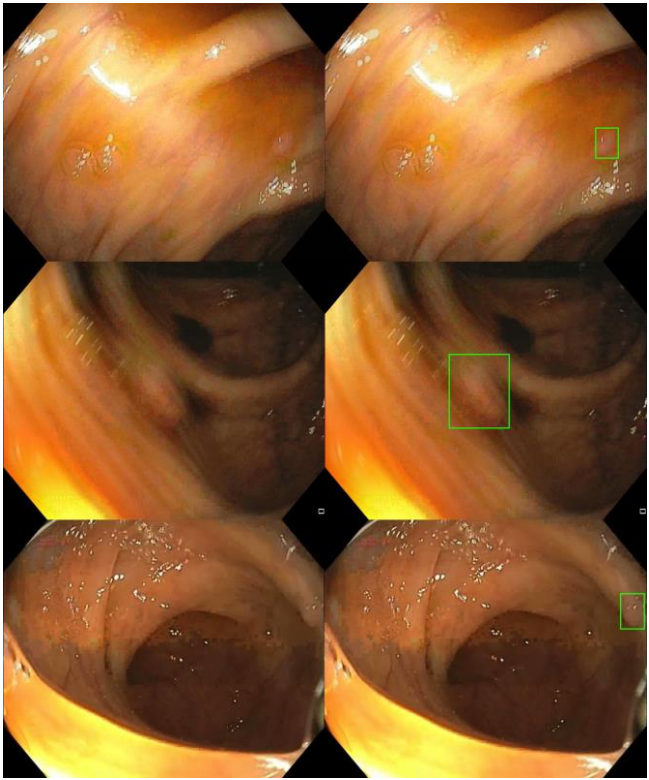

**FIGURE 9.** Examples of polyp image frames which were missed by experts for polyp masking (first column) yet correctly detected by our trained model (second column).

Another reason for the high number of FPs might be because the poor colon preparation before the colonoscopy examination in the AUS-MAYO video dataset. Therefore, there are many normal frames with many polyp-like objects, see Fig. 4. However, our model is trained using only polyp image frames; consequently, our model has not learned to distinguish polyp-like objects from actual polyps which lead to many FPs. As expected, after applying the automatic FP learning process (second row in Table IV), many of the FPs decrease, resulting in increased precision compared to when the FP learning results (Aug-1) are not used.

Table VI shows the results of the proposed automatic FP learning scheme on the ASU-Mayo 5 negative test videos which have 6854 normal frames. After applying the automatic FP learning, specificity improves by 26.6%, proving that the proposed FP learning scheme can efficiently decrease polyp-like FPs. We, therefore propose, that if the detection model is only trained with positive training samples, then the FP learning scheme will be a good tool for reliable detection systems.

TABLE VI
POLYP DETECTION RESULTS FOR 5 NEGATIVE VIDEOS (EACH VIDEO HAS NO POLYP FRAME)

| Method | Total frames | FP | TN | Spe (%) |
|---|---|---|---|---|
| Augmentation-I | 6854 | 1979 | 4875 | 71.1 |
| Automatic FP learning | 6854 | 161 | 6693 | **97.7** |

In Table V, even though the three models show the same 100% PDR and similar RT, after applying the FP learning scheme, the FPs are significantly decrease and results in improved precision by 14.7% compared to without FP learning (Aug-I). In addition, after applying the off-line learning method, we obtain better detection performance for all metrics compared to Aug-I and obtain better recall, F1 and F2 scores than with the FP learning scheme.

### E. EVALUATION OF COLONOSCOPY VIDEOS-II

For more reliable evaluation of the proposed detection system in video detection and to compare it with other methods, we include the new and larger public video database, i.e., CVC-ClinicVideoDB (see Section III for detailed database information).

In Table VII, we evaluate the effect of the different augmentation strategies on 18 test videos. The results in Table VII are highly consistent with the results of still frame dataset (Table I) and 10 test video dataset (Table IV). Thus, large training samples obtained by Aug-I and –II show much better performance compared to the small number of training samples. However, Aug-II (the largest training samples) shows no better performance than Aug-I in terms of Recall, F1 and F2 scores. Therefore, we conclude that the image augmentation has an important role to improve detection performance. However, as shown in Fig. 6, applying many different augmentations such as blurring and brightening to obtain a large number of training samples does not guarantee better detection performance, and we recommend that domain-specific characteristics have to be considered before applying augmentations.

TABLE VII
COMPARISON OF POLYP DETECTION RESULTS USING FOUR DIFFERENT AUGMENTATION STRATEGIES ON CVC-CLINICVIDEODB (18 VIDEOS)

| Train Dataset | TP | FP | FN | TN | Pre (%) | Rec (%) | F1 (%) | F2 (%) |
|---|---|---|---|---|---|---|---|---|
| w/o aug | 4308 | 2962 | 5717 | 1365 | 59.3 | 48 | 49.8 | 45.5 |
| Rot-aug | 6113 | 2981 | 3912 | 1143 | 67.2 | 61 | 64 | 62.1 |
| Aug-I | 8036 | 1645 | 1985 | 1151 | 83 | **80.2** | **81.6** | **80.7** |
| Aug-II | 7021 | 1079 | 3004 | 1509 | **86.7** | 70 | 77.5 | 72.8 |

Table VIII lists the results of our model based on Aug-I dataset, FP learning and off-line learning frameworks. Similar to the 10 video results in Table V, the off-line learning can improve the overall performance of the model based on Aug-I, and the FP learning can considerably decrease the number of FPs and leads to the best precision in Table VIII.

In Table VIII, we compare our results to the results in [31], where the studies used exactly the same training and testing datasets. In [31], it was suggested to use the AdaBoost learning strategy to train an initial classifier based on image patch based feature types such as LBP and/or Haar. In their post learning process, they re-train the initial classifier using the new selected negative examples (FPs). In the last two rows of Table VIII, Ni (e.g., HaarN1) refers to a classifier computed with i-th re-training steps.

The results in Table VIII indicate that our models show better performance compared to their results regarding all metrics except the mean processing time per frame (MPT). The big difference in performance improvement, i.e., recall, precision, F1- and F2-score, might be due to the use of the deep CNN model instead of the hand-craft features in [31]. In our model, the MPT highly depends on the hardware systems, i.e., GPU and the CNN architectures. Even though we did not optimize the method to improve the detection time, as shown in Section IV-B, the MPT of the proposed system (390ms) is competitive with other CNN based polyp detection methods. In the future, we aim to optimize the network architecture in conjunction with the GPU class to speed up the detection time.

## V. CONCLUSION

We present a deep learning based automatic polyp detection system in this study. A Faster R-CNN method incorporated with a recent deep-CNN model, Inception Resnet, is adopted for this detection system. The main benefit of the proposed system is the superior detection performance in terms of precision, recall and reaction time (RT) in both image and video databases. Furthermore, the proposed detector system is simply trained using whole image frames instead of conventional patch extraction (polyp and background) based training. Due to the use of very deep CNN in our detector system, the detection processing time in each frame is about 0.39 sec. This might be a disadvantage of the system and should be improved in the future if real-time detection is required as in standard colonoscopy. For WCE detection systems where off-line detection is more acceptable, the time delay may be of less importance.

## ACKNOWLEDGMENT

The authors would like to express sincere appreciation to Jennifer Morrison at OmniVision Technologies Norway AS for her valuable comments.

TABLE VIII
POLYP DETECTION RESULTS FOR 18 POSITIVE VIDEOS (EACH VIDEO HAS AT LEAST ONE POLYP FRAME)

| Method | Pre (%) | Rec (%) | F1 (%) | F2 (%) | MPT (msec) | PDR | RT (frames, sec) |
|---|---|---|---|---|---|---|---|
| Aug-I | 83 | 80.2 | 81.6 | 80.3 | 390 | 100% | 1.61, 0.064 |
| FP learning | **92.2** | 69.7 | 79.4 | 73.3 | 390 | 100% | 12,9, 0.51 |
| Off-line learning | 89.7 | **84.3** | **86.9** | **85.3** | 390 | 100% | **1.5, 0.06** |
| HaarN1 [31] | 39.1 | 42.6 | 40.8 | 41.8 | **21** | 100% | 27.3, 1.1 |
| LBPN2+HaarN1[31] | 30.4 | 52.4 | 38.5 | 45.8 | 185 | 100% | 15.0, 0.6 |

## REFERENCES

[1] R. L. Siegel, K. D. Miller and A. Jemal, "Cancer statistics 2017," *CA Cancer J Clin.,* vol. 67, pp. 7-30, 2017.
[2] M. Gschwantler, S. Kriwanek, E. Langner, B. Göritzer, C. Schrutka-Kölbl, E. Brownstone, H. Feichtinger and W. Weiss, "High-grade dysplasia and invasive carcinoma in colorectal adenomas: a multivariate analysis of the impact of adenoma and patient characteristics," *Eur. J. Gastroenterol. Hepatol.*, vol. 14, no. 2, pp. 183–188, 2002.
[3] A. M. Leufkens, M. G. H. van Oijen, F. P. Vleggaar and P. D. Siersema, "Factors influencing the miss rate of polyps in a back-to-back colonoscopy study," *Endoscopy*, vol. 44, no. 5, pp. 470–475, 2012.
[4] L. Rabeneck, J. Souchek and H. B. El-Serag, "Survival of colorectal cancer patients hospitalized in the Veterans Affairs Health Care System," *Am J Gastroenterol.*, vol. 98, no. 5, pp. 1186-1192, 2003.
[5] S. A. Karkanis, D. K. Iakovidis, D. E. Maroulis, D. A. Karras, and M. Tzivras, "Computer-aided tumor detection in endoscopic video using color wavelet features," *IEEE Trans. Inf. Technol. Biomed.*, vol. 7, no.

3, pp. 141–152, 2003.
[6] S. Ameling, S. Wirth, D. Paulus, G. Lacey and F. Vilario, "Texture-based polyp detection in colonoscopy" in Bildverarbeitung fr die Medizin 2009, Germany, Berlin:Springer, pp. 346-350, 2009.
[7] S. Hwang, J. Oh, W. Tavanapong, J. Wong, and P. de Groen, "Polyp detection in colonoscopy video using elliptical shape feature," in *Proc. IEEE Int. Conf. Image Process.*, vol. 2, pp. II-465-468, 2007.
[8] J. Bernal, J. Snchez, and F. Vilarino, "Towards automatic polyp detection with a polyp appearance model," *Pattern Recognit.*, vol. 45, no. 9, pp. 3166–3182, 2012.
[9] J. Bernal, J. Sánchez, and F. Vilarino, "Impact of image preprocessing methods on polyp localization in colonoscopy frames," in *Proc. 35th Annu. Int. Conf. IEEE EMBC*, pp. 7350–7354, 2013.
[10] J. Bernal, J. Snchez, G. F.-Esparrach, D. Gil, C. Rodriguez and F. Vilario, "Wm-dova maps for accurate polyp highlighting in colonoscopy: Validation vs. saliency maps from physicians," *Comput. Med. Imag. Graph.*, vol. 43, pp. 99–111, 2015.
[11] N. Tajbakhsh, S. Gurudu, and J. Liang, "A classification-enhanced vote accumulation scheme for detecting colonic polyps," in Abdominal Imaging. Computation and Clinical Applications, New York:Springer, vol. 8198, pp. 53-62, 2013.
[12] N. Tajbakhsh, S. R. Gurudu and J. Liang, "Automated Polyp Detection in Colonoscopy Videos Using Shape and Context Information," *IEEE Trans. Med. Imag.*, vol. 35, no.2, pp. 630-644, 2016.
[13] S. Park, M. Lee, and N. Kwak, "Polyp detection in colonoscopy videos using deeply-learned hierarchical features," *Seoul Nat. Univ.*, 2015.
[14] S. Park and D. Sargent, "Colonoscopic polyp detection using convolutional neural networks," *SPIE Med. Imag.*, p. 978528, 2016.
[15] N. Tajbakhsh, J. Y. Shin, S. R. Gurudu, R. T. Hurst, C. B. Kendall, M. B. Gotway and Jianming Liang, "Convolutional neural networks for medical image analysis: Full training or fine tuning?" *IEEE Trans. Med. Imag.*, vol. 35, no. 5, pp. 1299–1312, May 2016.
[16] L. Yu, H. Chen, Q. Dou, J. Qin and P. A. Heng, "Integrating online and offline 3D deep learning for automated polyp detection in colonoscopy videos," *IEEE J. Biomed. Health Inform.*, vol. 21, no.1, pp.65-75, 2017
[17] S. Bae and K. Yoon, "Polyp detection via imbalanced learning and discriminative feature learning", IEEE Trans. Med. Imag., vol. 34, no. 11, pp. 2379-2393, 2015.
[18] R. Girshick, J. Donahue, T. Darrell, and J. Malik, "Rich feature hierarchies for accurate object detection and semantic segmentation," in *Proc. IEEE Conf. on Computer Vision and Pattern Recognition (CVPR)*, Columbus, OH, pp. 580–587, 2014.
[19] R. Girshick, "Fast R-CNN," in *Proc. IEEE Int. Conf. on Computer Vision*, Santiago, Chile, pp. 1440–1448, 2015.
[20] S. Ren, K. He, R. Girshick, and J. Sun, "R-CNN: Towards real-time object detection with region proposal networks," *in Advances in Neural Information Processing Systems*, Montreal, QC, pp. 91–99, 2015.
[21] J. R. R. Uijlings, K. E. A. van de Sande, T. Gevers, and A. W. M. Smeulders, "Selective search for object recognition," *Int. J. Comput. Vis.*, vol. 104, no. 2, pp. 154–171, 2013.
[22] C. L. Zitnick and P. Dollar, "Edge boxes: Locating object proposals from edges," in *Proc. Computer Vision - ECCV 2014*, Springer Lecture Notes in Computer Science, vol. 8963, pp. 391–405, 2015.
[23] K. He, G. Gkioxari, P. Dollar, and R. Girshick., "Mask R-CNN", *arXiv preprint arXiv:1703.06870*, 2017.
[24] L. Zhang, L. Lin, X. Liang, and K. He. "Is faster r-cnn doing well for pedestrian detection?," *in European Conference on Computer Vision (ECCV)*, pp. 443-457, 2016.
[25] X. Zhao, W. Li, Y. Zhang, T. A. Gulliver, S. Chang and Z. Feng, "A Faster RCNN-Based Pedestrian Detection System," *IEEE 84th Vehicular Technology Conference (VTC-Fall)*, pp. 1-5, 2016
[26] H. Jiang and E. Learned-Miller, "Face detection with the faster r-cnn", arXiv preprint arXiv:1606.03473, 2016.
[27] J. Bernal, N. Tajbkaksh,, F. J. Sánchez, J. Matuszewski, H. Chen, L. Yu, Q. Angermann, O. Romain, B. Rustad, I. Balasingham, K. Pogorelov, S. Choi, Q. Debard, L. M. Hen, S. Speidel, D. Stoyanov, P. Brandao, H. Cordova, C. S. Montes, S. R. Gurudu, G. F. Esparrach, X. Dray, J. Liang and A. Histace, "Comparative Validation of Polyp Detection Methods in Video Colonoscopy: Results from the MICCAI 2015 Endoscopic Vision Challenge," *IEEE Trans. Med. Imaging*, vol. 36, no. 6, pp. 1231-49, 2017.
[28] H. Chen et al., "Standard plane localization in fetal ultrasound via domain transferred deep neural networks," IEEE J. Biomed. Health Informat., vol. 19, no. 5, pp. 1627–1636, Sep. 2015.
[29] H. Shin, L. Lu, L. Kim, A. Seff, J. Yao, and R. Summers, "Interleaved text/image deep mining on a large-scale radiology image database," in IEEE Conf. on CVPR, 2015, pp. 1–10
[30] H.-C. Shin, H. R. Roth, M. Gao, L. Lu, Z. Xu, I. Nogues, J. Yao, D. Mollura and R. M. Summers, "Deep convolutional neural networks for computer-aided detection: CNN architectures dataset characteristics and transfer learning", IEEE Trans. Med. Imag., vol. 35, no. 5, pp. 1285-1298, 2016.
[31] Q. Angermann, J. Bernal, C. Sánchez-Montes, M. Hammami, G. Fernández-Esparrach, X. Dray, O. Romain, F. J. Sánchez and A. Histace, "Towards Real-Time Polyp Detection in Colonoscopy Videos: Adapting Still Frame-Based Methodologies for Video Sequences Analysis" In Computer Assisted and Robotic Endoscopy and Clinical Image-Based Procedures, Springer, Cham., pp. 29-41, 2017.
[32] Alex Krizhevsky, Ilya Sutskever, and Geoffrey E Hinton. "Imagenet classification with deep convolutional neural networks", In Neural Information Processing Systems (NIPS), 2012.
[33] J. Huang, V. Rathod, C. Sun, M. Zhu, A. Korattikara, A. Fathi, I. Fischer, Z. Wojna, Y. Song, S. Guadarrama and K. Murphy, "Speed/accuracy trade-offs for modern convolutional object detectors", in Proc. IEEE Conf. on Computer Vision and Pattern Recognition (CVPR), 2017.
[34] X. Chen and A. Gupta, "An Implementation of Faster RCNN with Study for Region Sampling", arXiv preprint arXiv:1702.02138, 2017.
[35] C. Szegedy, S. Ioffe, and V. Vanhoucke, "Inception-v4, inception-resnet and the impact of residual connections on learning", arXiv:1602.07261, 2016.
[36] K. He, X. Zhang, S. Ren, and J. Sun, "Deep residual learning for image recognition," arXiv:1512.03385, 2015.
[37] C. Szegedy, V. Vanhoucke, S. Ioffe, J. Shlens, and Z. Wojna, "Rethinking the inception architecture for computer vision", arXiv:1512.00567, 2015
[38] C. Szegedy, W. Liu, Y. Jia, P. Sermanet, S. Reed, D. Anguelov, D. Erhan, V. Vanhoucke, and A. Rabinovich, "Going deeper with convolutions", in Proc. IEEE Conf. on Computer Vision and Pattern Recognition (CVPR), 2015.
[39] https://github.com/tensorflow/models/blob/master/research/slim/nets/inception_resnet_v2.py
[40] T.-Y. Lin, M. Maire, S. Belongie, J. Hays, P. Perona, D. Ramanan, P. Dollar, and C. L. Zitnick. "Microsoft COCO: Common objects in context", in European Conference on Computer Vision (ECCV), 2014.
[41] J. S. Silva, A. Histace, O. Romain, X. Dray and B. Granado, "Towards embedded detection of polyps in WCE images for early diagnosis of colorectal cancer," *Int J Comput Assist Radiol Surg.*, vol. 9, no. 2, pp. 283-293, 2014.
[42] H. Chen, X. J. Qi, J. Z. Cheng, and P. A. Heng, "Deep contextual networks for neuronal structure segmentation," *in Thirtieth AAAI Conference on Artificial Intelligence*, 2016.